\documentclass[lettersize,journal]{IEEEtran}
\usepackage{amsmath,amsfonts}
\usepackage{algorithmic}
\usepackage{algorithm}
\usepackage{array}
\usepackage[caption=false,font=normalsize,labelfont=sf,textfont=sf]{subfig}
\usepackage{textcomp}
\usepackage{stfloats}
\usepackage{url}
\usepackage{verbatim}
\usepackage{graphicx}
\usepackage{cite}
\usepackage{multirow}
\usepackage[table,xcdraw]{xcolor}
\usepackage[capitalize]{cleveref}
\usepackage{booktabs}
\usepackage{pifont}

\begin{document}

\title{BPG: Balancing Plasticity and Generalization for Domain Incremental Learning}

\author{Qiang~Wang,
Songlin~Dong,
Shaokun~Wang,
Jizhou~Han,
Xiang~Song,
Chenhao~Ding,
Yuhang~He,
and~Yihong~Gong,~\IEEEmembership{Fellow,~IEEE}

\thanks{Qiang~Wang, Jizhou~Han, Yuhang~He, and Yihong~Gong are with the College of Artificial Intelligence, Xi'an Jiaotong University, Xi'an, China. 
Songlin~Dong is with Shenzhen University of Advanced Technology, Shenzhen, China. 
Shaokun~Wang is with Harbin Institute of Technology, Shenzhen, China. 
Xiang~Song and Chenhao~Ding are with the School of Software Engineering, Xi'an Jiaotong University, Xi'an, China.
}

}

\markboth{Preprint}%
{Wang \MakeLowercase{\textit{et al.}}: BPG: Balancing Plasticity and Generalization for DIL}

\maketitle

\begin{abstract}
Deep neural networks excel in various tasks but struggle to generalize across evolving data distributions, leading to significant performance degradation under domain shifts. Domain incremental learning (DIL) addresses this challenge by enabling models to continuously adapt while retaining prior knowledge. Among existing DIL approaches, the parameter-isolation paradigm achieves state-of-the-art performance. However, these methods often adopt a one-size-fits-all approach to adapt to new domains, resulting in either insufficient learning capacity or redundant parameters. In this work, we propose \textbf{BPG}, a unified framework that addresses both challenges through two complementary components: \textbf{BPG-Adapter}, which dynamically determines each domain's adapter hidden dimension based on domain-specific feature separability, and \textbf{BPG-Inference}, a soft domain mixture strategy that integrates multiple domain-specific models at test time, mitigating domain ID misselection. Experimental results on DomainNet, CDDB, and CORe50 demonstrate that BPG consistently outperforms uniform adapter-based approaches and hard domain selection strategies, achieving state-of-the-art average accuracy while reducing forgetting to as low as 0.22\% on DomainNet.
\end{abstract}


\section{Introduction}
\IEEEPARstart{D}{eep} neural networks (DNNs) have achieved remarkable success in visual recognition tasks such as image classification~\cite{zhang2024query,li2024dynamic,ding2025class,han2025learn,lu2026gfpl}, object detection~\cite{chen2018robust,song2025learning,zhao2026shared,shi2026decenter}, and semantic segmentation~\cite{li2025towards,tang2023holistic}. However, these models typically assume that training and test data share the same distribution, making them prone to catastrophic forgetting~\cite{mccloskey1989catastrophic,ding2024lobg,han2026goal} in non-stationary environments where lighting conditions~\cite{li2026trajectory}, sensor characteristics, or visual styles change~\cite{han2025consistent,dong2025beyond} over time. This is especially concerning in safety-critical applications such as autonomous driving~\cite{yang2024preventing}, medical imaging~\cite{perkonigg2021dynamic}, and surveillance systems~\cite{doshi2020continual,wang2025vdc}.

\textbf{Domain Incremental Learning (DIL)}~\cite{sprompt,tan2020incremental,gao2024beyond,hu2025video,tao2024class,tpdiod,inflora,li2025dc} addresses this challenge: a model is sequentially exposed to domains that exhibit distinct visual characteristics, and must incorporate new domain knowledge while preserving performance on all previously learned domains. Existing DIL approaches fall into three paradigms: rehearsal-based methods~\cite{isele2018selective,rolnick2019experience,zhao2021memory} that replay stored past samples but incur storage and privacy costs, regularization-based methods~\cite{ewc,zenke2017continual,aljundi2018memory,akyurek2021subspace,liu2022few,shi2023multi} that constrain parameter updates to protect old knowledge but limit the absorption of new knowledge, and parameter-isolation methods~\cite{sprompt,mopclip,pina,cprompt,wang2024importance} that freeze a shared backbone and learn lightweight domain-specific modules, achieving state-of-the-art performance by explicitly preventing cross-domain interference. While parameter-isolation methods have shown strong results, they struggle to balance the fundamental tension between \emph{plasticity} (the ability to learn effectively from new domains) and \emph{generalization} (the capacity to maintain robust performance across all encountered domains).

\begin{figure*}[t]
  \centering
  \includegraphics[width=0.7\textwidth]{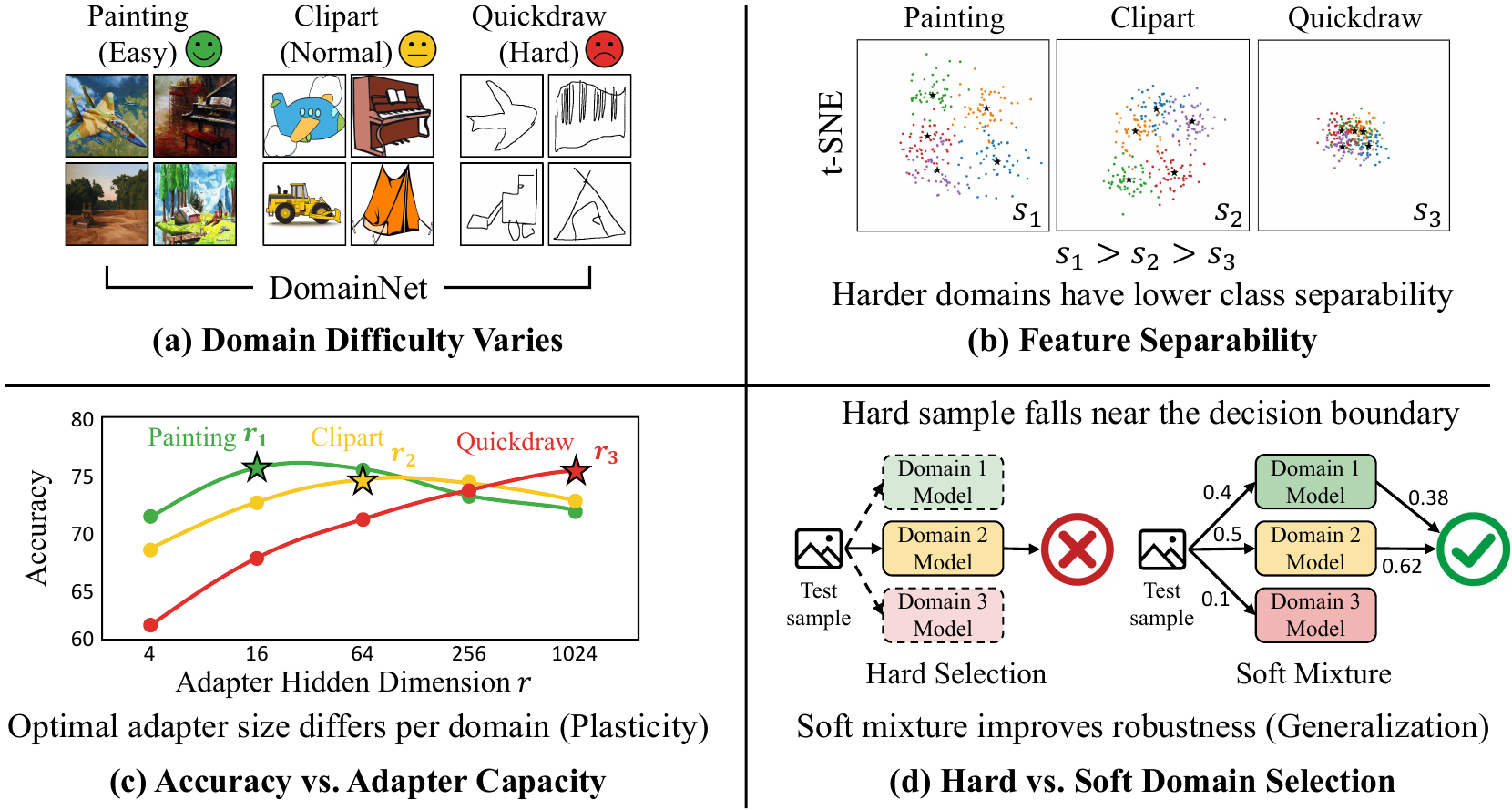}
\vspace{-3mm}
\caption{\textbf{Motivation of BPG.}
    (a) Domains from DomainNet exhibit varying levels of visual complexity.
    (b) t-SNE visualizations of pre-trained backbone (ViT-B/16) features confirm that harder domains have lower feature separability scores ($s_1 > s_2 > s_3$).
    (c) The optimal adapter dimension differs across domains: easier domains peak at smaller capacities while harder domains benefit from larger ones, motivating our adaptive allocation strategy (BPG-Adapter).
    (d) Hard domain selection at inference can misassign ambiguous samples near decision boundaries, whereas our soft mixture strategy (BPG-Inference) aggregates predictions from multiple domain-specific models weighted by confidence, improving robustness.}
    \label{fig:intro_fig}
\vspace{-3mm}
\end{figure*}

\textbf{The Plasticity Challenge: One Size Does Not Fit All.} Most existing methods assign fixed-capacity modules (e.g., uniform adapters or prompts) to every domain, overlooking the substantial variation in domain difficulty. As illustrated in~\cref{fig:intro_fig}~(a-b), applying a pre-trained backbone to three DomainNet domains yields markedly different feature-space separability, ranging from well-separated clusters in the Painting domain to heavily entangled ones in the Quickdraw domain. This disparity has direct implications for model design. As shown in~\cref{fig:intro_fig}~(c), easier domains attain peak performance at small adapter dimensions and may even degrade when granted excessive capacity through overfitting, whereas harder domains continue to benefit from larger capacity. Consequently, a uniform adapter design is suboptimal: allocating too few parameters to a hard domain undermines plasticity, while allocating too many to an easy domain induces overfitting and impairs generalization.

\textbf{The Generalization Challenge: Hard Selection is Brittle.} Maintaining separate parameters per domain requires a domain identification step at inference: given a test sample, the model selects a single domain ID and applies only the corresponding parameters. This hard selection is inherently fragile. As illustrated in~\cref{fig:intro_fig}~(d), samples near domain boundaries are prone to misassignment, causing a catastrophic mismatch between the test sample and the applied domain-specific model. The problem worsens as more domains are learned, and our analysis shows that domain ID errors in hard selection can degrade early-domain accuracy by over seven points.

We propose \textbf{BPG}, a unified framework that addresses both challenges through complementary mechanisms.
For plasticity, we introduce \textbf{BPG-Adapter}, an adaptive capacity allocation strategy that matches adapter parameters to each domain's intrinsic difficulty: domains already well separated by the pre-trained backbone need only lightweight adaptation, while domains with entangled representations require richer parameterization. Guided by a qualitative risk decomposition, BPG-Adapter operationalizes this insight through a simple inverse-proportionality rule that directs capacity where it is most needed.
For generalization, we introduce \textbf{BPG-Inference}, a soft domain mixture strategy that replaces hard selection with confidence-weighted aggregation over multiple domain-specific models. Rather than committing to a single domain, it estimates a distribution over all learned domains for each test sample, prunes unlikely candidates, and fuses the surviving predictions, drawing on complementary domain knowledge when a sample is ambiguous. Together, these components reduce average forgetting to as low as 0.22\% on DomainNet while attaining state-of-the-art average accuracy.
The main contributions are summarized below.
\begin{itemize}
    \item We identify the plasticity-generalization tension in parameter-isolation DIL and propose BPG, a unified framework that addresses both challenges through complementary mechanisms.
    \item We introduce BPG-Adapter, which dynamically allocates adapter capacity by feature separability, supported by a qualitative risk decomposition that motivates the inverse-proportionality allocation rule.
    \item We present BPG-Inference, a soft domain mixture strategy that mitigates the fragility of hard domain ID selection and serves as a plug-and-play module for other parameter-isolation methods.
    \item Extensive experiments on three multi-domain benchmarks (DomainNet, CDDB, and CORe50) show that BPG consistently achieves state-of-the-art performance in both average accuracy and forgetting rate.
\end{itemize}

\section{Related Work}
\subsection{Domain Incremental Learning}
Domain Incremental Learning (DIL) methods fall into three paradigms: rehearsal-based, regularization-based, and parameter-isolation.

\emph{Rehearsal-based} methods replay stored or generated past samples during subsequent training, via exemplar selection~\cite{kim2020imbalanced,ding2025space}, prototype preservation~\cite{zhang2019variational}, or generative replay~\cite{shin2017continual,van2021class}. On frozen pre-trained models, these ideas resurface in a data-free form for class-incremental learning: RanPAC~\cite{ranpac} and APER~\cite{aper} build prototype-based classifiers over frozen or first-session-adapted features, while SLCA~\cite{slca} models each class as a Gaussian and replays sampled pseudo-features to debias the classifier. However, their performance degrades as the exemplar budget shrinks~\cite{buzzega2020dark,li2026parameter}, and storing raw data raises privacy concerns~\cite{verma2023privacy}.

\emph{Regularization-based} methods add loss terms that constrain parameter updates to protect learned representations, e.g., EWC~\cite{ewc} penalizes changes to parameters important for earlier domains and distillation variants~\cite{shi2023multi} enforce feature-space consistency. For vision-language models, ZSCL~\cite{zscl} follows the same regularization philosophy, distilling against the initial CLIP model and averaging weights to prevent zero-shot transfer degradation. Although they avoid data storage, the finite capacity of a single model makes them overly conservative, struggling to absorb substantially different domains while retaining old knowledge~\cite{van2024continual}.

\emph{Parameter-isolation} methods instead maintain domain-specific parameters to prevent cross-domain interference. S-Prompts~\cite{sprompt} learns per-domain prompts on a frozen transformer and selects domains by K-NN matching over K-Means centroids, and MoP-CLIP~\cite{mopclip} extends this to CLIP-based prompt mixtures. Later works differ mainly in the isolated module: PINA~\cite{pina} adds a per-domain alignment module with a Patch Shuffle Selector, C-Prompt~\cite{cprompt} pools learnable prompts for cross-domain compositionality, ISPSL~\cite{wang2024importance} decouples domain-specific and domain-shared low-rank subspaces, and KA-Prompt~\cite{kaprompt} aligns knowledge across prompts component-wise. Pre-trained-model methods for class-incremental learning follow the same spirit with per-task modules, e.g., LAE~\cite{lae} ensembles online and offline expert adapters, and EASE~\cite{ease} learns a per-task adapter to form expandable subspaces. Despite strong performance, two limitations persist: a uniform-capacity module is attached to every domain, ignoring the wide variation in domain difficulty; and hard domain selection at inference is fragile near domain boundaries. Our BPG tackles both by allocating adapter capacity according to domain difficulty and replacing hard selection with a soft mixture over domain-specific experts.

\vspace{-2mm}
\subsection{Adaptive Parameter-Efficient Fine-Tuning}
Parameter-efficient fine-tuning (PEFT) methods, such as adapters~\cite{adapter} and LoRA~\cite{lora}, adapt pre-trained models by updating only a small fraction of parameters, but typically fix the rank or hidden dimension across all weight matrices and tasks. Since performance is highly sensitive to this choice~\cite{adapternlp}, a uniform budget overlooks that different matrices and tasks demand varying adaptation capacity. To address this, recent works adaptively allocate capacity, e.g., by pruning or budgeting ranks according to importance or intrinsic dimensionality~\cite{zhang2023adalora,valipour2022dylora,ed2024gelora,shinwari2025ard}, or by gating and selecting adapters at the layer, token, or module level~\cite{guo2021adaptive,zhou2024dynamic,qi2025adaptive}. However, these methods target single- or multi-task fine-tuning and allocate capacity across layers within a task; none explicitly links adapter capacity to the intrinsic difficulty of the target distribution, leaving open how to determine the right capacity per domain in incremental learning. Our BPG-Adapter fills this gap with a data-driven separability metric that guides hidden-dimension allocation across domains, motivated by a qualitative risk decomposition.

\vspace{-2mm}
\subsection{Test-Time Domain Identification}
Parameter-isolation methods for DIL must decide, at inference, which domain-specific parameters to activate for a given test sample, and the design of this mechanism critically affects performance. Most existing methods rely on \emph{hard} selection. S-Prompts~\cite{sprompt} stores per-domain K-Means centroids during training and assigns a single domain label via K-NN matching in the frozen feature space. Later methods refine this selection: PINA~\cite{pina} introduces a Patch Shuffle Selector that suppresses class-dependent cues and outperforms K-NN and Nearest Mean classifiers, while ESN~\cite{wang2023isolation} selects the most confident domain via a temperature-controlled energy metric over stage classifiers. C-Prompt~\cite{cprompt} instead pools and recomposes prompts across domains, implicitly softening the domain boundary. Nevertheless, committing to a single domain remains fragile for samples near domain boundaries.

A related paradigm is Mixture-of-Experts (MoE), which softly routes inputs across expert sub-networks via a learned gating network. In continual learning, Lifelong-MoE~\cite{chen2023lifelong} adds experts for emerging distributions while regularizing experts and gates to retain old knowledge, and MoE-Adapters~\cite{yu2024boosting,yu2025moe} treat adapters as experts within CLIP, using a Distribution Discriminative Auto-Selector to route between adapted and zero-shot models. Although these methods confirm the benefit of soft routing over hard selection, they train the gating mechanism jointly with the experts, incurring extra training cost and risking forgetting in the router itself.

In contrast, our BPG-Inference is a training-free, plug-and-play soft domain mixture. For each test sample, it estimates a confidence-weighted distribution over all learned domains, prunes unlikely candidates, and fuses predictions from the surviving domain-specific models. This retains the robustness of soft routing without any additional trainable module or test-time parameter update, and can be applied on top of existing parameter-isolation DIL methods.

\vspace{-2mm}
\section{Method}

\begin{figure*}[t]
  \centering
  \includegraphics[width=0.85\textwidth]{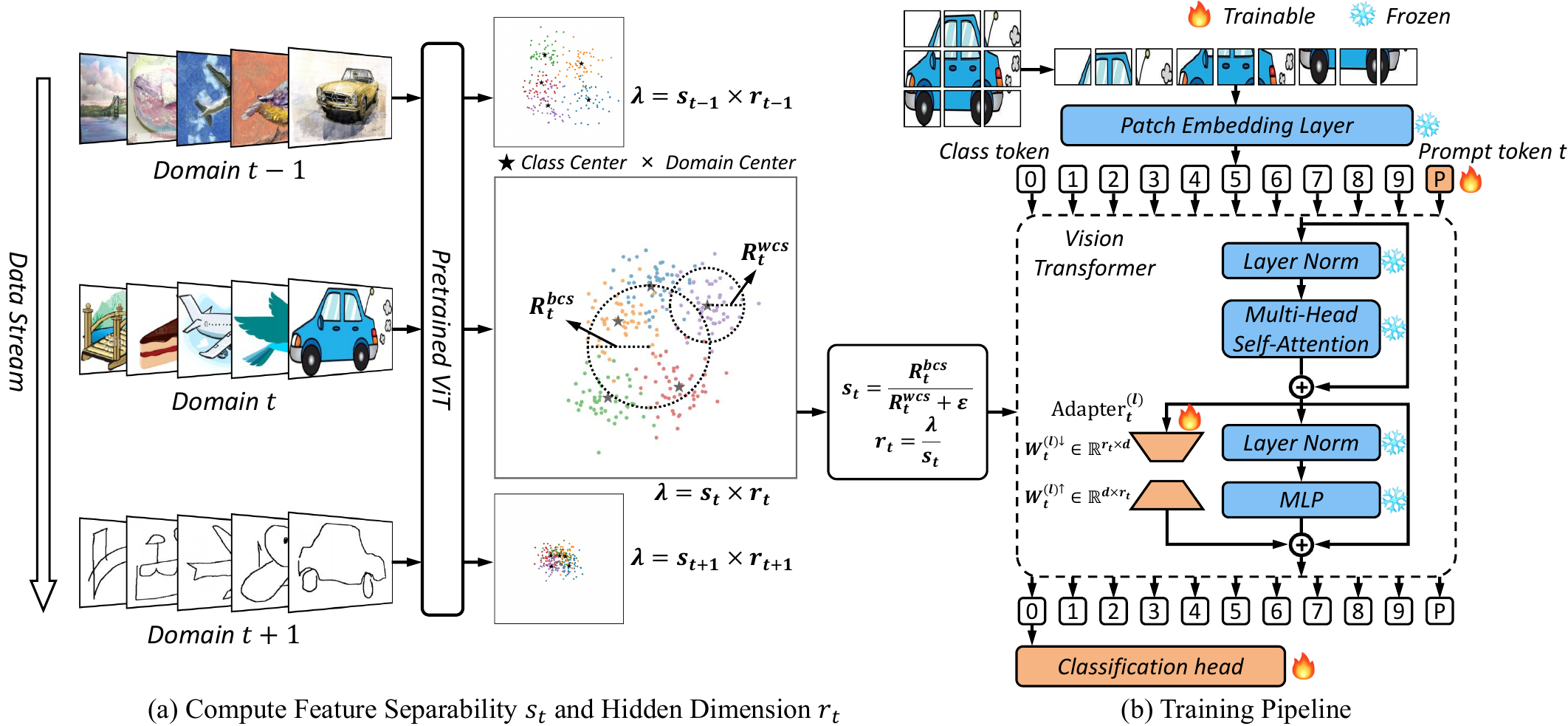}
  \vspace{-3mm}
  \caption{Overall architecture of the proposed BPG framework.
  An input image is tokenized by the patch embedding layer and concatenated with trainable prompt tokens and a class token.
  The token sequence is processed by a stack of frozen vision transformer layers equipped with domain-specific adapters.
  For each domain $t$, the adapter hidden dimension $r_t$ is determined by the feature separability score $s_t$.}
  \label{fig:framework}
  \vspace{-3mm}
\end{figure*}

\subsection{Problem Formulation}
\label{sec:problem_formulation}
We consider a domain incremental learning scenario with $T$ domains, where the dataset is written as $\mathcal{D}=\{\mathcal{D}_t\}_{t=1}^{T}$. Each domain dataset $\mathcal{D}_t=(\mathcal{X}_t,\mathcal{Z}_t)$ consists of a training set $\mathcal{X}_t$ and a test set $\mathcal{Z}_t$. Following the DIL protocol, during the $t$-th training session only samples in $\mathcal{X}_t$ are accessible, while data from other domains is not stored or replayed. After completing session $t$, the model is evaluated on previously seen test sets
\begin{equation}
  \mathcal{Z}_{\leq t} = \mathcal{Z}_{1} \cup \mathcal{Z}_{2} \cup \cdots \cup \mathcal{Z}_{t},
\end{equation}
and the performance across sessions is summarized by standard metrics, i.e., average accuracy and forgetting. The details of metrics will be provided in \cref{sec:exp_setting}.

\vspace{-3mm}
\subsection{Vanilla Adapter for Domain-Specific Knowledge}
\label{sec:vanilla_adapter}
Before introducing BPG-Adapter, we describe the vanilla adapter-based architecture that serves as the building block for our framework. This subsection explains how domain-specific adapters are inserted into a ViT encoder and how they interact with prompt tokens and classification heads.

\paragraph{Tokenization and Encoder Backbone}
Let $x$ denote an input image from domain $t$. $x$ is first divided into non-overlapping image patches, which are linearly projected to patch tokens and added with positional embeddings. We denote the resulting sequence of $N$ patch tokens as $\mathbf{x} = [\mathbf{x}_1,\ldots,\mathbf{x}_N] \in \mathbb{R}^{N \times d}$, where $d$ is the embedding dimension.

For domain $t$, we introduce domain-specific prompt tokens $\mathbf{p}_t \in \mathbb{R}^{M \times d}$ that are prepended to the patch tokens and a class token $\mathbf{x}_{\mathrm{cls}}\in\mathbb{R}^{1 \times d}$ that aggregates image-level information. The input sequence to the encoder at layer $\ell=0$ is
\begin{equation}
  \mathbf{z}^{(0)}_t = 
  [\mathbf{x}_{\mathrm{cls}},\, \mathbf{x}_1,\ldots,\mathbf{x}_N,\, \mathbf{p}_t]
  \in \mathbb{R}^{(N+M+1)\times d}.
\end{equation}
We employ a stack of $L$ transformer encoder layers with frozen parameters $\theta$. Each layer consists of a multi-head self-attention (MHSA) block and a feed-forward multi-layer perceptron (MLP) block with residual connections and layer normalization (LN), as shown in \cref{fig:framework}~(b). Let $\mathbf{z}^{(\ell)}_t$ be the input to layer $\ell$. The standard ViT update is
\begin{align}
  \mathbf{h}^{(\ell)}_t &= 
  \mathbf{z}^{(\ell)}_t 
  + \mathrm{MHSA}\big(\mathrm{LN}(\mathbf{z}^{(\ell)}_t);\theta_{\mathrm{att}}^{(\ell)}\big), \label{eq:vit_attn} \\
  \mathbf{z}^{(\ell+1)}_t &= 
  \mathbf{h}^{(\ell)}_t 
  + \mathrm{MLP}\big(\mathrm{LN}(\mathbf{h}^{(\ell)}_t);\theta_{\mathrm{mlp}}^{(\ell)}\big), \label{eq:vit_mlp}
\end{align}
where $\theta_{\mathrm{att}}^{(\ell)}$ and $\theta_{\mathrm{mlp}}^{(\ell)}$ are frozen pre-trained parameters.

\paragraph{Adapter Structure}
To capture domain-specific characteristics with a small number of trainable parameters, we insert an adapter module as an additional residual branch in each encoder layer, running in parallel to the feed-forward MLP block. For domain $t$ and layer $\ell$, the adapter is a two-layer bottleneck MLP
\begin{equation}
  \mathrm{Adapter}_t^{(\ell)}(\mathbf{u}) =
  \sigma\!\left(
  \mathbf{u}\,\mathbf{W}^{(\ell)\downarrow\top}_t
  \right)
  \mathbf{W}^{(\ell)\uparrow\top}_t, \label{eq:adapter_def}
\end{equation}
where $\mathbf{u}\in\mathbb{R}^{(N+M+1)\times d}$ stacks all token embeddings as rows (shared with the MLP branch), $\mathbf{W}^{(\ell)\downarrow}_t \in \mathbb{R}^{r_t \times d}$ and $\mathbf{W}^{(\ell)\uparrow}_t \in \mathbb{R}^{d \times r_t}$ are the down- and up-projection matrices, $r_t$ is the adapter hidden dimension of domain $t$, and $\sigma(\cdot)$ is a non-linear activation function (ReLU). Each matrix multiplication is applied independently to every token row, yielding an output in $\mathbb{R}^{(N+M+1)\times d}$. Because the adapter output is added to the same residual path as the MLP output and the adapter is initialized to produce near-zero outputs, the overall layer recovers the frozen pre-trained parameters at initialization, which stabilizes training.

Concretely, for domain $t$, the attention sub-layer in \eqref{eq:vit_attn} is unchanged. We then apply layer normalization and branch into two parallel paths: a frozen MLP branch and a trainable adapter branch:
\begin{align}
  \mathbf{m}^{(\ell)}_t &=
  \mathrm{MLP}\big(\mathrm{LN}(\mathbf{h}^{(\ell)}_t);\theta_{\mathrm{mlp}}^{(\ell)}\big), \\
  \mathbf{a}^{(\ell)}_t &=
  \mathrm{Adapter}_t^{(\ell)}\!\left(
  \mathrm{LN}(\mathbf{h}^{(\ell)}_t)
  \right).
\end{align}
The output of layer $\ell$ is obtained by adding the two branches to the residual connection:
\begin{equation}
  \mathbf{z}^{(\ell+1)}_t =
  \mathbf{h}^{(\ell)}_t
  + \mathbf{m}^{(\ell)}_t
  + \mathbf{a}^{(\ell)}_t .
  \label{eq:vit_layer_adapter}
\end{equation}
When the adapter branch is disabled ($\mathbf{a}^{(\ell)}_t=\mathbf{0}$), \eqref{eq:vit_layer_adapter} reduces to the standard ViT update in \eqref{eq:vit_mlp}.
All backbone parameters $\theta=\{\theta_{\mathrm{att}}^{(\ell)},\theta_{\mathrm{mlp}}^{(\ell)}\}_{\ell=1}^{L}$ are frozen, whereas adapter parameters $\{\mathbf{W}^{(\ell)\downarrow}_t,\mathbf{W}^{(\ell)\uparrow}_t\}_{\ell=1}^{L}$ and the prompt token $\mathbf{p}_t$ are trainable for domain $t$. This parallel-residual design creates an isolated parameter set per domain while fully reusing the same pre-trained transformer layer.

\paragraph{Domain-Specific Classification Head}
After passing through $L$ encoder layers, we obtain the final token sequence $\mathbf{z}^{(L)}_t$. We use the output of the class token (the first token in the sequence) as the image representation, denoted as $\mathbf{h}_t = \mathbf{z}^{(L)}_t[0] \in \mathbb{R}^{d}$. Each domain $t$ is associated with its own linear classifier
\begin{equation}
  \mathbf{y}_t = g_t(\mathbf{h}_t;\eta_t) = \mathbf{W}_t \mathbf{h}_t + \mathbf{b}_t,
  \label{eq:domain_logits}
\end{equation}
where $\eta_t=\{\mathbf{W}_t,\mathbf{b}_t\}$ and $\mathbf{y}_t \in \mathbb{R}^{C}$ are the class logits. For domain $t$, only $\mathbf{p}_t$, the adapter parameters and classifier $\eta_t$ are updated, while parameters from previous domains remain fixed. In the next subsections, we describe how BPG allocates the adapter capacity across domains and how it performs soft inference at test time.

\vspace{-3mm}
\subsection{BPG-Adapter: Domain-wise Capacity Allocation}
\label{sec:bpg_adapter}
The vanilla adapter in \cref{sec:vanilla_adapter} uses the same hidden dimension for all domains, i.e., $\forall t \in [1,T], r_t = r$. However, domains may exhibit different levels of difficulty: some align well with the pre-trained backbone, while others contain more confusing or fine-grained categories. Allocating an equal number of parameters to all domains may thus fail to model hard domains and over-parameterize easy ones. BPG-Adapter addresses this issue by dynamically determining the hidden dimension $r_t$ for each domain according to a \emph{feature separability} score.

\paragraph{Domain-wise Feature Separability}
Let $t\in\{1,\ldots,T\}$ index domains, $c\in\{1,\ldots,C\}$ index classes, and $i\in\{1,\ldots,N_{t,c}\}$ index training samples from class $c$ in domain $t$. We use a pre-trained feature extractor to extract a $d$-dimensional feature vector $\mathbf{x}_{t,c,i}\in\mathbb{R}^d$ for each sample. The class mean and domain mean are
\begin{equation}
  \boldsymbol{\mu}_{t,c} = 
  \frac{1}{N_{t,c}}\sum_{i=1}^{N_{t,c}}\mathbf{x}_{t,c,i},
  \quad
  \boldsymbol{\mu}_t =
  \frac{1}{C}\sum_{c=1}^{C}\boldsymbol{\mu}_{t,c}.
\end{equation}

We define the \emph{between-class scatter} (BCS) and \emph{within-class scatter} (WCS) of domain $t$ as
\begin{align}
  R^{\mathrm{bcs}}_t &= 
  \frac{1}{C}\sum_{c=1}^{C}
  \big\|
  \boldsymbol{\mu}_{t,c} - \boldsymbol{\mu}_t
  \big\|_2^2, \label{eq:bcs} \\
  R^{\mathrm{wcs}}_t &= 
  \frac{1}{C}\sum_{c=1}^{C}
  \frac{1}{N_{t,c}}
  \sum_{i=1}^{N_{t,c}}
  \big\|
  \mathbf{x}_{t,c,i} - \boldsymbol{\mu}_{t,c}
  \big\|_2^2, \label{eq:wcs}
\end{align}
where $\|\cdot\|_2$ is the Euclidean norm. Intuitively, a large $R^{\mathrm{bcs}}_t$ and a small $R^{\mathrm{wcs}}_t$ indicate that classes in domain $t$ are well separated and compact, which is favorable for classification. A clear example is shown in \cref{fig:framework}~(a). Then we combine these quantities into a scalar feature separability score
\begin{equation}
  s_t = 
  \frac{R^{\mathrm{bcs}}_t}
       {R^{\mathrm{wcs}}_t + \varepsilon},
  \label{eq:separability}
\end{equation}
where $\varepsilon=10^{-6}$ prevents division by zero. A higher $s_t$ implies better separability and thus an easier domain.

\paragraph{Capacity Allocation Rule}
To balance plasticity and generalization, we assume that each domain is allocated a certain amount of trainable capacity through its adapter hidden dimension $r_t$. We desire an inverse relationship between $s_t$ and $r_t$: hard domains (low $s_t$) should receive more parameters to improve plasticity, while easy domains (high $s_t$) can be modeled with smaller adapters to avoid unnecessary capacity. The detailed justification and derivation are provided in \cref{sec:theory}. We impose a simple multiplicative constraint $\lambda = s_t r_t$, $t=1,\ldots,T$,
where $\lambda$ is a constant controlling the overall capacity budget. In practice, instead of tuning $\lambda$ directly, we use a reference domain with its separability $s_0$ and a reference hidden dimension $r_0$, and set $\lambda = s_0 r_0$. Substituting into this constraint yields
\begin{equation}
  r_t = \frac{s_0}{s_t} r_0.
  \label{eq:rt_rule}
\end{equation}
We treat the ImageNet dataset~\cite{imagenet} as the reference domain because it is the pre-training dataset of the backbone and typically exhibits relatively high separability. By using \cref{eq:rt_rule}, BPG-Adapter automatically assigns more trainable parameters to challenging domains and less to easier ones, without manual per-domain tuning.

\vspace{-3mm}
\subsection{BPG-Inference: Soft Mixture for More Generalization}
\label{sec:bpg_inference}
The previous subsections describe how BPG learns a collection of domain-specific adapters and classifiers. At inference time, a crucial question is how to leverage these domain-specific models for an unseen test sample. Most existing methods perform \emph{hard} domain selection: a domain-ID predictor is first used to choose one domain, and only the corresponding adapter and classifier are applied. This strategy is brittle: when the predicted domain is wrong or when the sample lies near the boundary of multiple domains, the final prediction can be severely degraded. BPG-Inference replaces hard selection with a \emph{soft domain mixture}. Each domain-specific model is treated as an expert, and their logits are combined according to similarities between the test sample and each domain.

\paragraph{Prototype-based Domain Affinity}
For each domain $t$, we run $k$-means clustering ($k=5$) on the frozen feature representations of its training samples (the same features used in \cref{eq:bcs,eq:wcs}). This yields $k$ prototypes $\{\boldsymbol{\nu}_{t,j}\}_{j=1}^{k}$ that summarize the feature distribution of domain $t$. Given a test image, we extract its feature vector $\mathbf{z}\in\mathbb{R}^d$ using the frozen backbone and compute the distance to domain $t$ as the nearest-prototype distance
\begin{equation}
  a_t = \min_{j\in\{1,\ldots,k\}}
  \big\|
  \mathbf{z} - \boldsymbol{\nu}_{t,j}
  \big\|_2.
  \label{eq:nearest_distance}
\end{equation}
Smaller $a_t$ indicates that the sample is closer to domain $t$ in the feature space and thus more likely to be well handled by its adapter and classifier.

We convert distances into normalized domain confidences via
\begin{equation}
  w_t = 
  \frac{\exp(-a_t)}
       {\sum_{i=1}^{T}\exp(-a_i)},
  \quad t=1,\ldots,T.
  \label{eq:softmax_confidence}
\end{equation}
This softmax over negative distances assigns larger weights to domains whose prototypes are closer to the test sample.

\paragraph{Confidence-Guided Sparse Domain Mixture}
To suppress noisy contributions from irrelevant domains, we apply a parameter-free pruning step on the confidence scores $\{w_t\}_{t=1}^{T}$. We treat the uniform distribution over domains as a threshold: domains whose confidence is lower than this threshold are considered unlikely to be helpful for the current sample and their weights are set to zero. The remaining domains keep their original confidences, which are then rescaled as $w'_t$ so that the surviving weights sum to one. In this way, each test sample is associated with a small set of high-confidence domains whose classifiers are linearly combined, using the rescaled confidences as mixture weights, to produce the final logits, i.e., $\mathbf{y}_{\mathrm{final}} = \sum_{t=1}^{T} w'_t \mathbf{y}_t$. The final prediction is obtained by applying the softmax function to $\mathbf{y}_{\mathrm{final}}$. This soft domain mixture allows multiple domains to jointly contribute to decision making when appropriate and reduces the risk of catastrophic errors caused by incorrect hard domain selection.

\vspace{-3mm}
\subsection{Theoretical Insight for Capacity Allocation}
\label{sec:theory}
We now provide a qualitative theoretical motivation for the capacity allocation rule in \cref{eq:rt_rule} and discuss how it helps balance plasticity and generalization.
The goal is not to derive a tight bound or claim global optimality; rather, we use this analysis as intuitive guidance for why assigning larger adapters to harder domains and smaller adapters to easier domains is a sensible design choice.

\paragraph{A Toy Risk Decomposition}
Consider the expected classification risk on domain $t$ as a function of its adapter dimension $r_t$:
\begin{equation}
  \mathcal{R}_t(r_t) =
  \underbrace{\mathcal{R}^{\mathrm{approx}}_t(r_t)}_{\text{approximation error}}
  +
  \underbrace{\mathcal{R}^{\mathrm{gen}}_t(r_t)}_{\text{estimation error}}.
  \label{eq:risk_decomposition}
\end{equation}
The approximation error $\mathcal{R}^{\mathrm{approx}}_t(r_t)$ decreases as $r_t$ increases, because a larger adapter offers higher expressive power to fit domain-specific variations.
The estimation error $\mathcal{R}^{\mathrm{gen}}_t(r_t)$, on the other hand, typically increases with $r_t$, since a more complex model tends to have larger capacity and thus higher generalization error on finite samples~\cite{shalev2014understanding,mohri2018foundations}. For analytical convenience, we adopt the following simple parametric forms:
\begin{align}
  \mathcal{R}^{\mathrm{approx}}_t(r_t)
  &\approx A_t \exp(-\kappa_t r_t), \label{eq:approx_error} \\
  \mathcal{R}^{\mathrm{gen}}_t(r_t)
  &\approx B \sqrt{\frac{r_t}{N_t}}, \label{eq:gen_error}
\end{align}
where $A_t > 0$ and $\kappa_t > 0$ describe how fast the approximation error decays with $r_t$ for domain $t$, $B>0$ is a constant, and $N_t$ is the number of training samples.
We emphasize that these two expressions are illustrative surrogates rather than exact laws: the exponential form captures the diminishing returns of adding capacity, echoing the geometric error decay of expressive models in classical approximation arguments, while the $\sqrt{r_t/N_t}$ term mirrors standard capacity-based generalization bounds~\cite{shalev2014understanding,mohri2018foundations}. Importantly, the derivation that follows uses only their qualitative shape, namely an approximation error that decreases in $r_t$ and an estimation error that increases in $r_t$, so the same conclusion holds for any pair of terms sharing this monotone structure.
The parameter $\kappa_t$ reflects domain difficulty: when classes are well separated (large $s_t$), the error can be reduced quickly with small $r_t$ (larger $\kappa_t$); when classes are highly entangled (small $s_t$), more parameters are needed, corresponding to smaller $\kappa_t$.

\paragraph{Qualitative Allocation Under a Capacity Budget}
Suppose that the total adapter capacity is constrained by a budget $\sum_{t=1}^{T} r_t \leq R_{\mathrm{tot}}$.
We approximate the average risk by
\begin{equation}
  \bar{\mathcal{R}} =
  \frac{1}{T}\sum_{t=1}^{T}
  \mathcal{R}_t(r_t).
\end{equation}
Minimizing $\bar{\mathcal{R}}$ under the budget constraint can be approached via the Lagrangian
\begin{equation}
  \mathcal{L} =
  \frac{1}{T}\sum_{t=1}^{T}
  \Big(A_t e^{-\kappa_t r_t} + B \sqrt{\tfrac{r_t}{N_t}} \Big)
  + \alpha \Big(\sum_{t=1}^{T} r_t - R_{\mathrm{tot}}\Big),
\end{equation}
where $\alpha \geq 0$ is a Lagrange multiplier.
We differentiate $\mathcal{L}$ with respect to $r_t$ and set $\frac{\partial \mathcal{L}}{\partial r_t} = 0$, which yields the stationarity condition
\begin{equation}
  \underbrace{A_t \kappa_t e^{-\kappa_t r_t^\star}}_{\text{marginal plasticity gain}}
  =
  \underbrace{\frac{B}{2\sqrt{N_t r_t^\star}} + \alpha T}_{\text{marginal generalization / budget cost}}.
  \label{eq:stationarity}
\end{equation}
\cref{eq:stationarity} is precisely the plasticity--generalization balance we seek: at the optimum, the marginal reduction of approximation error obtained by enlarging the adapter (left-hand side, \emph{plasticity}) is equated with the marginal increase of estimation error together with the shadow price of the capacity budget (right-hand side, \emph{generalization}). The allocation is thus determined by the equality of the two competing terms, rather than by discarding either of them.
Although \cref{eq:stationarity} admits no elementary closed-form solution, its structure is informative. The left-hand side decays exponentially in $r_t$, whereas the right-hand side, $g_t(r_t) := \frac{B}{2\sqrt{N_t r_t}} + \alpha T$, varies only slowly (algebraically) with $r_t$. Solving for $r_t^\star$ therefore gives
\begin{equation}
  r_t^\star \approx
  \frac{1}{\kappa_t}
  \ln\!\left(
  \frac{A_t \kappa_t}{\,g_t(r_t^\star)\,}
  \right),
  \label{eq:r_star}
\end{equation}
where $\ln$ denotes the natural logarithm.
Because the generalization/budget cost $g_t$ enters only through the logarithm, it acts as a slowly varying factor and is retained rather than neglected: a stronger generalization penalty (larger $B$ or smaller sample size $N_t$) or a tighter budget (larger $\alpha$) uniformly shrinks every $r_t^\star$, while the \emph{relative} allocation across domains is governed by the fast exponential term. Consequently, $r_t^\star$ scales roughly like $1/\kappa_t$ up to this slowly varying logarithmic factor: harder domains (smaller $\kappa_t$) are assigned larger adapter dimensions, whereas easier domains (larger $\kappa_t$) can be modeled with smaller adapters.

The remaining step is to connect the abstract decay rate $\kappa_t$ to the observable separability score $s_t$. Crucially, obtaining the desired trend does \emph{not} require an exact functional relationship: since $r_t^\star$ in \cref{eq:r_star} decreases with $\kappa_t$, \emph{any} monotonically increasing map $\kappa_t = \varphi(s_t)$ already yields $r_t^\star$ decreasing in $s_t$, i.e., more separable (easier) domains receive smaller adapters. This monotone relationship is corroborated by the empirical behavior reported in \cref{sec:visualization}: the least separable domains (e.g., Infograph, $s_2{=}0.124$, and Quickdraw, $s_4{=}0.151$) keep benefiting from larger $r$ up to $r{=}1024$, whereas the most separable one (Real, $s_5{=}0.530$) already peaks at a small $r$ and degrades under excess capacity; uniformly enlarging $r$ therefore helps hard domains at the cost of more severe degradation on easier ones, indicating a faster error decay (larger $\kappa_t$) for easier domains. Adopting the simplest such map, $\kappa_t \propto s_t$, \cref{eq:r_star} then reduces to the clean trend
\begin{equation}
  r_t^\star \propto \frac{1}{s_t},
\end{equation}
which is consistent with our capacity allocation rule in \cref{eq:rt_rule}. 
Therefore, BPG-Adapter can be viewed as a practical heuristic that approximates an optimal allocation of limited capacity across domains with different difficulties.

\paragraph{Implications for Plasticity and Generalization}
The analysis above highlights the trade-off between plasticity and generalization.
Allocating too few parameters to a hard domain results in high approximation error and thus poor plasticity; allocating too many parameters to an easy domain inflates the total capacity, which can harm global generalization under a fixed budget.
By adapting $r_t$ according to $1/s_t$, BPG-Adapter increases plasticity where it is most needed and saves parameters where the backbone already captures the domain well.
Together with BPG-Inference, which promotes cross-domain sharing at test time via soft mixtures, the BPG framework achieves a more balanced trade-off between learning new domains and preserving performance on past ones.

\section{Experiments}
\subsection{Experimental Settings}
\label{sec:exp_setting}
\paragraph{Datasets}
To evaluate our method, we conduct experiments on three widely used multi-domain datasets: DomainNet, CDDB, and CORe50. DomainNet~\cite{domainnet} is a large-scale dataset designed for domain adaptation and domain incremental learning. It consists of six distinct domains with significant inter-domain variations, each containing 345 categories. The training set comprises 409,832 images, while the test set includes 176,743 images. CDDB~\cite{cddb} is tailored for continual deepfake detection, encompassing multiple deepfake techniques. Following previous works, we adopt the Hard track, which includes GauGAN, BigGAN, WildDeepfake, WhichFaceReal, and SAN as representative deepfake generation methods. CORe50~\cite{core50} is an object recognition dataset structured into 11 domains, with eight designated for training and three for testing. Each domain contains 50 classes and approximately 15,000 images.

\paragraph{Evaluation Metrics}
To assess the performance of DIL methods, we utilize two key evaluation metrics. We denote by $B \in \mathbb{R}^{T \times T}$ a lower triangular matrix, where $B_{i,j}$ is the test accuracy on the $j$-th domain after training on the $i$-th domain. (1) Average Accuracy ($A_{T}$) measures the classification performance after training on the $T$-th domain; it is computed as the overall accuracy over all test images from the $T$ learned domains. (2) Average Forgetting ($F_{T}$), which measures the retention of knowledge from previous domains, is computed as:
\begin{equation}
\label{eq:FT}
  F_{T} = \frac{1}{T-1} \sum_{j=1}^{T-1} \frac{1}{T-j} \sum_{i=j+1}^{T}(B_{j,j}-B_{i,j}).
\end{equation}
A higher $A_T$ and a lower $F_T$ indicate better performance of a DIL method.

\paragraph{Backbones}
Following~\cite{sprompt}, we adopt pre-trained ViT-B/16~\cite{vit} and CLIP~\cite{clip} as backbones. ViT extracts image features with a classifier trained from scratch; CLIP additionally encodes class-name text features and uses image-text similarity as classification logits.

\paragraph{Implementation Details}
We train the model for 30, 50, and 20 epochs on the DomainNet, CDDB, and CORe50 datasets, respectively. The model is optimized using SGD with an initial learning rate of 0.01, following a cosine decay schedule. The batch size is set to 128, and all training images are resized to $224 \times 224$. Each experiment is repeated three times with different random seeds, and we report the mean and standard deviation. The between-class scatter ($R^{bcs}_{t}$), within-class scatter ($R^{wcs}_{t}$), and feature separability score ($s_{t}$) for each domain in these datasets are presented in \cref{tab:separability_domainnet,tab:separability_cddb,tab:separability_core50}, where $t=0$ denotes ImageNet as the reference domain for computing $r_t$. The reference hidden dimension $r_{0}$ is set to 64 for DomainNet and CORe50, and to 1 for CDDB. All experiments are conducted on a workstation with an Intel Xeon Gold 6226R CPU, 320\,GB RAM, and NVIDIA RTX 4090 GPUs; each run requires a single GPU with approximately 18--22\,GB of GPU memory.

\begin{table}[t]
\centering
\caption{Between-class scatter $R^{bcs}_t$, within-class scatter $R^{wcs}_t$, and feature separability score $s_t$ on the DomainNet dataset.}
\label{tab:separability_domainnet}
\setlength{\tabcolsep}{4pt}
\vspace{-2mm}
\begin{tabular}{cccccccc}
\toprule
$t$           & 0     & 1     & 2     & 3     & 4     & 5     & 6     \\
\midrule
$R^{bcs}_{t}$ & 0.392 & 0.120 & 0.072 & 0.192 & 0.029 & 0.298 & 0.088 \\
$R^{wcs}_{t}$ & 0.518 & 0.543 & 0.579 & 0.656 & 0.193 & 0.562 & 0.568 \\
$s_{t}$       & 0.756 & 0.221 & 0.124 & 0.293 & 0.151 & 0.530 & 0.156 \\
\bottomrule
\end{tabular}
\vspace{-2mm}
\end{table}

\begin{table}[t]
\centering
\caption{Between-class scatter $R^{bcs}_t$, within-class scatter $R^{wcs}_t$, and feature separability score $s_t$ on the CDDB dataset.}
\label{tab:separability_cddb}
\setlength{\tabcolsep}{4pt}
\vspace{-2mm}
\begin{tabular}{ccccccc}
\toprule
$t$           & 0     & 1     & 2     & 3     & 4     & 5     \\
\midrule
$R^{bcs}_{t}$ & 0.392 & 0.007 & 0.004 & 0.002 & 0.004 & 0.009 \\
$R^{wcs}_{t}$ & 0.518 & 0.832 & 0.892 & 0.440 & 0.547 & 0.842 \\
$s_{t}$       & 0.756 & 0.008 & 0.004 & 0.005 & 0.008 & 0.011 \\
\bottomrule
\end{tabular}
\vspace{-2mm}
\end{table}

\begin{table}[t]
\centering
\caption{Between-class scatter $R^{bcs}_t$, within-class scatter $R^{wcs}_t$, and feature separability score $s_t$ on the CORe50 dataset.}
\label{tab:separability_core50}
\setlength{\tabcolsep}{2.3pt}
\vspace{-2mm}
\begin{tabular}{cccccccccc}
\toprule
$t$           & 0     & 1     & 2     & 3     & 4     & 5     & 6     & 7     & 8     \\
\midrule
$R^{bcs}_{t}$ & 0.392 & 0.205 & 0.208 & 0.178 & 0.162 & 0.219 & 0.167 & 0.199 & 0.185 \\
$R^{wcs}_{t}$ & 0.518 & 0.355 & 0.331 & 0.347 & 0.368 & 0.336 & 0.363 & 0.388 & 0.307 \\
$s_{t}$       & 0.756 & 0.579 & 0.629 & 0.513 & 0.441 & 0.650 & 0.460 & 0.513 & 0.604 \\
\bottomrule
\end{tabular}
\vspace{-2mm}
\end{table}

\begin{table}[t]
\setlength{\tabcolsep}{4pt}
\centering
\caption{\textbf{Experimental results on the DomainNet dataset.} ``Upper Bound'' denotes an ideal scenario where the domain ID of the test set is known. * Results reported in the original paper. \textbf{Bold}/\underline{underline}: best/second-best.}
\label{tab:domainnet}
\vspace{-2mm}
\begin{tabular}{clccc}
\toprule
Backbone              & Method                                & Buffer ($\downarrow$)     & $A_{T}$ ($\uparrow$)                 & $F_{T}$ ($\downarrow$)              \\
\midrule
\multirow{19}{*}{ViT} & DyTox~\cite{dytox}                    & 50/class                  & 62.94                                & -                                   \\
\cmidrule{2-5}
                      & EWC~\cite{ewc}                        & \multirow{16}{*}{0/class}                          & 47.62{\scriptsize$\pm$1.35}          & 12.85{\scriptsize$\pm$1.03}         \\
                      & LwF~\cite{lwf}                        &                           & 49.19{\scriptsize$\pm$1.13}          & 5.01{\scriptsize$\pm$0.59}          \\
                      & L2P~\cite{l2p}                        &                           & 40.15{\scriptsize$\pm$2.73}          & 2.25{\scriptsize$\pm$0.51}          \\
                      & DualPrompt~\cite{dualprompt}          &                           & 43.79{\scriptsize$\pm$1.55}          & 2.03{\scriptsize$\pm$0.89}          \\
                      & S-iPrompts~\cite{sprompt}             &                           & 50.62{\scriptsize$\pm$0.15}          & 2.85{\scriptsize$\pm$0.28}          \\
                      & CODA-P~\cite{coda}                    &                           & 47.42{\scriptsize$\pm$0.78}          & 3.46{\scriptsize$\pm$0.89}          \\
                      & PINA~\cite{pina}                      &                           & 54.86{\scriptsize$\pm$0.71}          & 2.24{\scriptsize$\pm$0.84}          \\
                      & DUCT*~\cite{duct}                      &                           & 67.01{\scriptsize$\pm$1.35}          & -                                   \\
                      & C-Prompt~\cite{cprompt}               &                           & 58.68{\scriptsize$\pm$1.28}          & 1.34{\scriptsize$\pm$0.36}          \\
                      & DualCP~\cite{dualcp}                  &                           & 60.13{\scriptsize$\pm$2.16}          & 1.96{\scriptsize$\pm$0.75}           \\
                      & SOYO~\cite{soyo}                      &                           & 65.25{\scriptsize$\pm$0.92}               & \underline{1.26{\scriptsize$\pm$0.41}} \\
                      & CONEC-LoRA*~\cite{conec}               &                           & 66.42{\scriptsize$\pm$0.38}          & -                                   \\
                      & KA-Prompt*~\cite{kaprompt}             &                           & 62.91{\scriptsize$\pm$0.14}          & 1.93{\scriptsize$\pm$0.26}         \\
                      & DCE~\cite{dce}                        &                           & 63.50{\scriptsize$\pm$0.50}          & -                                   \\
                      & PC~\cite{pc}                          &                           & 58.82{\scriptsize$\pm$0.46}          & 1.27{\scriptsize$\pm$0.18} \\
                      & ICON~\cite{icon}                      &                           & \underline{67.95{\scriptsize$\pm$1.87}} & 8.18{\scriptsize$\pm$1.80}          \\
\cmidrule{2-5}
                      & BPG (ours)                            & \multirow{2}{*}{0/class}  & \textbf{72.19{\scriptsize$\pm$0.86}} & \textbf{0.22{\scriptsize$\pm$0.12}} \\
                      & Upper Bound                           &                           & 74.21{\scriptsize$\pm$0.25}          & 0.00                                \\
\midrule
\multirow{7}{*}{CLIP} & S-liPrompts~\cite{sprompt}            & \multirow{5}{*}{0/class}  & 67.78{\scriptsize$\pm$0.90}          & 1.64{\scriptsize$\pm$0.68}          \\
                      & MoP-CLIP*~\cite{mopclip}               &                           & 69.70                                & -                                   \\
                      & PINA~\cite{pina}                      &                           & 69.06{\scriptsize$\pm$1.61}          & \underline{1.59{\scriptsize$\pm$0.33}} \\
                      & HiDe-Prompt*~\cite{hideprompt}         &                           & 60.15                                & -                                   \\
                      & CP-Prompt*~\cite{cpprompt}             &                           & \underline{73.35}                    & -                                   \\
\cmidrule{2-5}
                      & BPG (ours)                            & \multirow{2}{*}{0/class}  & \textbf{75.72{\scriptsize$\pm$0.15}} & \textbf{0.59{\scriptsize$\pm$0.03}} \\
                      & Upper Bound                           &                           & 77.53{\scriptsize$\pm$0.39}          & 0.00                                \\
\bottomrule
\end{tabular}
\vspace{-2mm}
\end{table}

\begin{table}[t]
\setlength{\tabcolsep}{4pt}
\centering
\caption{\textbf{Experimental results on the CDDB dataset.} ``Upper Bound'' denotes an ideal scenario where the domain ID of the test set is known. * Results reported in the original paper. \textbf{Bold}/\underline{underline}: best/second-best.}
\label{tab:cddb}
\vspace{-2mm}
\begin{tabular}{clccc}
\toprule
Backbone              & Method                        & Buffer ($\downarrow$)      & $A_{T}$ ($\uparrow$)                 & $F_{T}$ ($\downarrow$)              \\
\midrule
\multirow{23}{*}{ViT} & LRCIL~\cite{lrcil}            & \multirow{3}{*}{100/class} & 76.39                                & 4.39                                \\
                      & iCaRL~\cite{icarl}            &                            & 79.76                                & 8.73                                \\
                      & LUCIR~\cite{lucir}            &                            & 82.53                                & 5.34                                \\
\cmidrule{2-5}
                      & LRCIL~\cite{lrcil}            & \multirow{4}{*}{50/class}  & 74.01                                & 8.62                                \\
                      & iCaRL~\cite{icarl}            &                            & 73.98                                & 14.50                               \\
                      & LUCIR~\cite{lucir}            &                            & 80.77                                & 7.85                                \\
                      & DyTox~\cite{dytox}            &                            & 86.21                                & 1.55                                \\
\cmidrule{2-5}
                      & EWC~\cite{ewc}                & \multirow{14}{*}{0/class}  & 50.59{\scriptsize$\pm$3.79}          & 42.62{\scriptsize$\pm$1.83}         \\
                      & LwF~\cite{lwf}                &                            & 60.94{\scriptsize$\pm$1.69}          & 13.53{\scriptsize$\pm$0.50}         \\
                      & DyTox~\cite{dytox}            &                            & 51.27{\scriptsize$\pm$5.07}          & 45.85{\scriptsize$\pm$1.85}         \\
                      & L2P~\cite{l2p}                &                            & 61.28{\scriptsize$\pm$0.52}          & 9.23{\scriptsize$\pm$0.23}          \\
                      & DualPrompt~\cite{dualprompt}  &                            & 64.80{\scriptsize$\pm$1.82}          & 8.74{\scriptsize$\pm$0.73}          \\
                      & S-iPrompts~\cite{sprompt}     &                            & 74.51{\scriptsize$\pm$1.96}          & 1.30{\scriptsize$\pm$1.12}          \\
                      & CODA-P~\cite{coda}            &                            & 70.54{\scriptsize$\pm$0.37}          & 5.53{\scriptsize$\pm$0.94}          \\
                      & PINA~\cite{pina}              &                            & 77.35{\scriptsize$\pm$1.46}          & 0.98{\scriptsize$\pm$0.22} \\
                      & C-Prompt~\cite{cprompt}       &                            & 78.44{\scriptsize$\pm$0.87}          & 1.55{\scriptsize$\pm$0.15}      \\
                      & KA-Prompt~\cite{kaprompt}     &                            & 80.78{\scriptsize$\pm$1.21}          & 0.96{\scriptsize$\pm$0.26}                                \\
                      & DualCP~\cite{dualcp}          &                            & 82.16{\scriptsize$\pm$1.53}          & \underline{0.73{\scriptsize$\pm$0.25}}                                   \\
                      & CONEC-LoRA*~\cite{conec}       &                            & \underline{88.21{\scriptsize$\pm$0.88}}         & -                                   \\
                      & DUCT*~\cite{duct}              &                            & 85.10{\scriptsize$\pm$0.52}          & -                                   \\
                      & DCE*~\cite{dce}                &                            & 71.80{\scriptsize$\pm$4.20}          & -                                   \\
\cmidrule{2-5}
                      & BPG (ours)                    & \multirow{2}{*}{0/class}   & \textbf{88.55{\scriptsize$\pm$0.32}} & \textbf{0.68{\scriptsize$\pm$0.09}} \\
                      & Upper Bound                   &                            & 89.17{\scriptsize$\pm$0.40}          & 0.00                                \\
\midrule
\multirow{7}{*}{CLIP} & S-liPrompts~\cite{sprompt}    & \multirow{5}{*}{0/class}   & 88.65{\scriptsize$\pm$0.64}          & 0.69{\scriptsize$\pm$0.26}          \\
                      & MoP-CLIP*~\cite{mopclip}       &                            & 88.54                                & 0.79                                \\
                      & PINA~\cite{pina}              &                            & 85.71{\scriptsize$\pm$1.99}          & 0.51{\scriptsize$\pm$0.27}          \\
                      & HiDe-Prompt*~\cite{hideprompt} &                            & 84.32                                & 2.61                                \\
                      & CP-Prompt*~\cite{cpprompt}     &                            & \underline{93.65}                    & \underline{0.25}                    \\
\cmidrule{2-5}
                      & BPG (ours)                    & \multirow{2}{*}{0/class}   & \textbf{93.91{\scriptsize$\pm$0.39}} & \textbf{0.12{\scriptsize$\pm$0.06}} \\
                      & Upper Bound                   &                            & 94.40{\scriptsize$\pm$0.25}          & 0.00                                \\
\bottomrule
\end{tabular}
\vspace{-2mm}
\end{table}

\begin{table}[t]
\footnotesize
\setlength{\tabcolsep}{7pt}
\centering
\caption{\textbf{Experimental results on the CORe50 dataset.} Since the training and testing domains do not overlap, average forgetting and upper bound accuracy are not applicable. * Results reported in the original paper. \textbf{Bold}/\underline{underline}: best/second-best.}
\label{tab:core50}
\vspace{-2mm}
\begin{tabular}{clcc}
\toprule
Backbone              & Method                        & Buffer ($\downarrow$)     & $A_{T}$ ($\uparrow$)                 \\
\midrule
\multirow{24}{*}{ViT} & ER~\cite{er}                  & \multirow{7}{*}{50/class} & 80.10                                \\
                      & GDumb~\cite{gdumb}            &                           & 74.92                                \\
                      & BiC~\cite{bic}                &                           & 79.28                                \\
                      & DER++~\cite{der}              &                           & 79.70                                \\
                      & Co$^{2}$L~\cite{co2l}         &                           & 79.75                                \\
                      & DyTox~\cite{dytox}            &                           & 79.21                                \\
                      & L2P~\cite{l2p}                &                           & 81.07                                \\
\cmidrule{2-4}
                      & EWC~\cite{ewc}                & \multirow{15}{*}{0/class} & 74.82{\scriptsize$\pm$1.81}          \\
                      & LwF~\cite{lwf}                &                           & 75.45{\scriptsize$\pm$1.27}          \\
                      & L2P~\cite{l2p}                &                           & 78.33{\scriptsize$\pm$2.12}          \\
                      & DualPrompt~\cite{dualprompt}  &                           & 80.25{\scriptsize$\pm$1.09}          \\
                      & S-iPrompts~\cite{sprompt}     &                           & 83.13{\scriptsize$\pm$1.62}          \\
                      & ICON~\cite{icon}              &                           & 74.98{\scriptsize$\pm$0.03}          \\
                      & CODA-P~\cite{coda}            &                           & 85.68{\scriptsize$\pm$0.31}          \\
                      & C-Prompt~\cite{cprompt}       &                           & 85.31{\scriptsize$\pm$1.87}          \\
                      & KA-Prompt~\cite{kaprompt}     &                           & 85.61{\scriptsize$\pm$1.64}          \\
                      & PINA~\cite{pina}              &                           & 86.74{\scriptsize$\pm$0.82}          \\
                      & DCE*~\cite{dce}               &                           & 84.80{\scriptsize$\pm$0.30}          \\
                      & DualCP~\cite{dualcp}          &                           & 88.10{\scriptsize$\pm$0.89}          \\
                      & CONEC-LoRA*~\cite{conec}      &                           & 90.24{\scriptsize$\pm$1.64}          \\
                      & SOYO~\cite{soyo}              &                           & 90.77{\scriptsize$\pm$0.53}          \\
                      & PC~\cite{pc}                  &                           & \underline{91.35{\scriptsize$\pm$0.39}} \\
\cmidrule{2-4}
                      & BPG (ours)                    & \multirow{2}{*}{0/class}  & \textbf{91.87{\scriptsize$\pm$0.31}} \\
                      & Upper Bound                   &                           & N/A                                  \\
\midrule
\multirow{7}{*}{CLIP} & S-liPrompts~\cite{sprompt}    & \multirow{5}{*}{0/class}  & 89.06{\scriptsize$\pm$1.44}                       \\
                      & MoP-CLIP*~\cite{mopclip}       &                           & \underline{92.29}                    \\
                      & PINA~\cite{pina}              &                           & 87.38{\scriptsize$\pm$1.92}          \\
                      & HiDe-Prompt*~\cite{hideprompt} &                           & 80.81{\scriptsize$\pm$0.76}          \\
                      & CP-Prompt*~\cite{cpprompt}     &                           & 90.67{\scriptsize$\pm$0.55}          \\
\cmidrule{2-4}
                      & BPG (ours)                    & \multirow{2}{*}{0/class}  & \textbf{92.46{\scriptsize$\pm$0.42}} \\
                      & Upper Bound                   &                           & N/A                                  \\
\bottomrule
\end{tabular}
\vspace{-2mm}
\end{table}

\subsection{Main Results}

\paragraph{Compared Methods}
We compare BPG with representative continual learning methods across three paradigms: rehearsal-based (iCaRL~\cite{icarl}, LUCIR~\cite{lucir}, DyTox~\cite{dytox}), rehearsal-free regularization (EWC~\cite{ewc}, LwF~\cite{lwf}), and rehearsal-free parameter isolation, including prompt-based methods (L2P~\cite{l2p}, DualPrompt~\cite{dualprompt}, S-Prompts~\cite{sprompt}, CODA-P~\cite{coda}, C-Prompt~\cite{cprompt}) and recent approaches (ICON~\cite{icon}, DUCT~\cite{duct}, CONEC-LoRA~\cite{conec}, DCE~\cite{dce}, PC~\cite{pc}, PINA~\cite{pina}). On the CLIP backbone, we additionally include MoP-CLIP~\cite{mopclip}, HiDe-Prompt~\cite{hideprompt}, and CP-Prompt~\cite{cpprompt}. BPG is a rehearsal-free method that requires no replay buffer. We also report the Upper Bound accuracy, which assumes oracle domain IDs at test time, as a performance ceiling.

\paragraph{Results on DomainNet}
\cref{tab:domainnet} summarizes results on DomainNet, which comprises six visually distinct natural image domains. On ViT, BPG achieves $A_T{=}72.19\%$ and $F_T{=}0.22\%$, surpassing the second-best rehearsal-free method ICON~\cite{icon} ($67.95\%$) by 4.24\% in accuracy and SOYO~\cite{soyo} ($1.26\%$) by 1.04\% in forgetting. Without any replay, BPG also exceeds DyTox~\cite{dytox} (50 samples/class; $62.94\%$) by 9.25\%. The near-zero forgetting indicates that BPG-Inference mitigates errors from hard domain ID selection by softly combining domain-specific experts. BPG further narrows the gap to the Upper Bound ($74.21\%$) to 2.02\%. On CLIP, BPG attains $75.72\%$ and $0.59\%$ in $A_T$ and $F_T$, respectively, exceeding CP-Prompt~\cite{cpprompt} ($73.35\%$) by 2.37\% while maintaining low forgetting.

\paragraph{Results on CDDB}
\cref{tab:cddb} reports results on CDDB, a continual deepfake detection benchmark where each domain corresponds to a distinct generation method. On ViT, BPG achieves the best $A_T$ ($88.55\%$) and $F_T$ ($0.68\%$). Without storing past samples, it outperforms all rehearsal-based methods, including DyTox~\cite{dytox} (50 samples/class; $86.21\%$) and LUCIR~\cite{lucir} (100 samples/class; $82.53\%$). Its forgetting is 0.05\% lower than DualCP~\cite{dualcp} ($0.73\%$). On CLIP, BPG reaches $93.91\%$ with $0.12\%$ forgetting, surpassing CP-Prompt~\cite{cpprompt} ($93.65\%$) and approaching the Upper Bound ($94.40\%$) within 0.49\%.

\paragraph{Results on CORe50}
\cref{tab:core50} reports results on CORe50, where training and test domains do not overlap; consequently, $F_T$ and the Upper Bound are not applicable. BPG-Inference therefore softly aggregates predictions from all learned domain-specific models according to feature-space affinity, rather than committing to a single domain ID. On ViT, BPG attains $91.87\%$, outperforming the second-best method PC~\cite{pc} ($91.35\%$) and exceeding L2P~\cite{l2p} with a 50-sample buffer ($81.07\%$) by 10.80\%. On CLIP, BPG reaches $92.46\%$, surpassing MoP-CLIP~\cite{mopclip} ($92.29\%$) by 0.17\%. These results confirm that BPG generalizes effectively to entirely unseen test domains.

\subsection{Ablation Study}

\paragraph{Component-wise Analysis} \cref{tab:ab_component} presents the ablation results, where the baseline trains only the domain prompt and classifier on top of the frozen backbone and performs hard domain selection at inference. Both BPG-Adapter and BPG-Inference independently improve accuracy while reducing forgetting. By adaptively allocating larger capacity to harder domains, BPG-Adapter attains 68.33\% accuracy with 1.28\% forgetting on DomainNet (ViT), and BPG-Inference is particularly effective at reducing forgetting.

\paragraph{Generality of BPG-Inference} To demonstrate the broad applicability of our soft mixture strategy, we integrate BPG-Inference into several representative prompt-based continual learning methods, as shown in \cref{tab:ab_other_method}. BPG-Inference consistently improves all baseline methods across all three benchmarks. For instance, when applied to S-iPrompts, it yields +5.42\% accuracy gain on DomainNet with 0.74\% forgetting reduction. Similar improvements are observed for PINA (+4.48\% on DomainNet), C-Prompt (+4.83\% on DomainNet), and KA-Prompt (+2.52\% on DomainNet). These results confirm that BPG-Inference is a general and effective strategy that can enhance existing parameter-isolation methods by mitigating the brittleness of hard domain selection at inference time.

\begin{table*}[t]
\centering
\caption{Ablation study on BPG components. We evaluate the individual and combined contributions of BPG-Adapter (adaptive capacity allocation) and BPG-Inference (soft mixture strategy) using both ViT-B/16 and CLIP-ViT-B/16 backbones. ``Baseline'' trains only the domain prompt and classifier (without adapters) and uses hard domain selection at inference. Both components consistently improve performance, and their combination achieves the best results across all benchmarks.}
\label{tab:ab_component}
\vspace{-2mm}
\begin{tabular}{lccccccccc}
\toprule
\multirow{2}{*}{Method}  & \multicolumn{2}{c}{Backbone} & \multirow{2}{*}{BPG-Adapter} & \multirow{2}{*}{BPG-Inference} & \multicolumn{2}{c}{DomainNet}                                              & \multicolumn{2}{c}{CDDB}                                                   & CORe50                               \\
                         & ViT           & CLIP         &                              &                                & $A_{T}$ ($\uparrow$)                 & $F_{T}$ ($\downarrow$)              & $A_{T}$ ($\uparrow$)                 & $F_{T}$ ($\downarrow$)              & $A_{T}$ ($\uparrow$)                 \\
\midrule
Baseline                 & \ding{51}     &              &                              &                                & 61.69{\scriptsize$\pm$1.68}          & 1.86{\scriptsize$\pm$0.46}          & 80.52{\scriptsize$\pm$1.21}          & 0.93{\scriptsize$\pm$0.07}          & 83.23{\scriptsize$\pm$1.07}          \\
Baseline + BPG-Adapter   & \ding{51}     &              & \ding{51}                    &                                & 68.33{\scriptsize$\pm$0.89}          & 1.28{\scriptsize$\pm$0.17}          & 86.94{\scriptsize$\pm$1.93}          & 0.85{\scriptsize$\pm$0.42}          & 90.15{\scriptsize$\pm$1.09}          \\
Baseline + BPG-Inference & \ding{51}     &              &                              & \ding{51}                      & 66.25{\scriptsize$\pm$1.53}          & 1.54{\scriptsize$\pm$0.30}          & 83.84{\scriptsize$\pm$2.07}          & 0.77{\scriptsize$\pm$0.51}          & 84.55{\scriptsize$\pm$0.58}          \\
BPG (ours)               & \ding{51}     &              & \ding{51}                    & \ding{51}                      & \textbf{72.19{\scriptsize$\pm$0.86}} & \textbf{0.22{\scriptsize$\pm$0.12}} & \textbf{88.55{\scriptsize$\pm$0.32}} & \textbf{0.68{\scriptsize$\pm$0.09}} & \textbf{91.87{\scriptsize$\pm$0.31}} \\
\midrule
Baseline                 &               & \ding{51}    &                              &                                & 67.85{\scriptsize$\pm$0.72}          & 1.43{\scriptsize$\pm$0.26}          & 91.07{\scriptsize$\pm$1.49}          & 0.30{\scriptsize$\pm$0.44}          & 86.87{\scriptsize$\pm$1.12}          \\
Baseline + BPG-Adapter   &               & \ding{51}    & \ding{51}                    &                                & 73.59{\scriptsize$\pm$0.79}          & 1.15{\scriptsize$\pm$0.41}          & 93.42{\scriptsize$\pm$0.32}          & 0.26{\scriptsize$\pm$0.21}          & 92.09{\scriptsize$\pm$1.06}          \\
Baseline + BPG-Inference &               & \ding{51}    &                              & \ding{51}                      & 70.46{\scriptsize$\pm$0.47}          & 1.21{\scriptsize$\pm$0.23}          & 91.76{\scriptsize$\pm$1.38}          & 0.31{\scriptsize$\pm$0.29}          & 87.50{\scriptsize$\pm$1.91}          \\
BPG (ours)               &               & \ding{51}    & \ding{51}                    & \ding{51}                      & \textbf{75.72{\scriptsize$\pm$0.15}} & \textbf{0.59{\scriptsize$\pm$0.03}} & \textbf{93.91{\scriptsize$\pm$0.39}} & \textbf{0.12{\scriptsize$\pm$0.06}} & \textbf{92.46{\scriptsize$\pm$0.42}} \\
\bottomrule
\end{tabular}
\vspace{-2mm}
\end{table*}

\begin{table*}[t]
\centering
\caption{Generality of BPG-Inference. We integrate our soft mixture inference strategy into existing prompt-based continual learning methods. BPG-Inference consistently improves all baselines across benchmarks, demonstrating its broad applicability as a plug-and-play enhancement. Numbers in parentheses indicate improvements over the original methods.}
\label{tab:ab_other_method}
\vspace{-2mm}
\begin{tabular}{lccccc}
\toprule
\multirow{2}{*}{Method}     & \multicolumn{2}{c}{DomainNet}                 & \multicolumn{2}{c}{CDDB}                      & CORe50               \\
                            & $A_{T}$ ($\uparrow$) & $F_{T}$ ($\downarrow$) & $A_{T}$ ($\uparrow$) & $F_{T}$ ($\downarrow$) & $A_{T}$ ($\uparrow$) \\
\midrule
S-iPrompts~\cite{sprompt}   & 50.62                & 2.85                   & 74.51                & 1.30                   & 83.13                \\
\rowcolor{gray!20}
S-iPrompts + BPG-Inference & 56.04 (+5.42)        & 2.11 (-0.74)           & 77.69 (+3.18)        & 1.10 (-0.20)           & 83.80 (+0.67)         \\
PINA~\cite{pina}            & 54.86                & 2.24                   & 77.35                & 0.98                   & 86.74                \\
\rowcolor{gray!20}
PINA + BPG-Inference       & 59.34 (+4.48)        & 1.72 (-0.52)           & 79.49 (+2.14)        & 0.84 (-0.14)           & 87.35 (+0.61)        \\
C-Prompt~\cite{cprompt}     & 58.68                & 1.34                   & 78.44                & 1.55                   & 85.31                \\
\rowcolor{gray!20}
C-Prompt + BPG-Inference   & 63.51 (+4.83)        & 1.03 (-0.31)           & 80.57 (+2.13)        & 1.30 (-0.25)           & 86.45 (+1.14)        \\
KA-Prompt~\cite{kaprompt}   & 62.91                & 1.93                   & 80.78                & 0.96                   & 85.61                \\
\rowcolor{gray!20}
KA-Prompt + BPG-Inference  & 65.43 (+2.52)        & 1.19 (-0.74)           & 83.52 (+2.74)        & 0.75 (-0.21)           & 86.02 (+0.41)        \\
\bottomrule
\end{tabular}
\vspace{-2mm}
\end{table*}

\subsection{Additional Analysis on BPG-Adapter}
\label{sec:ana_bpg_adapter}

\begin{figure}[t]
  \centering
  \includegraphics[width=0.48\textwidth]{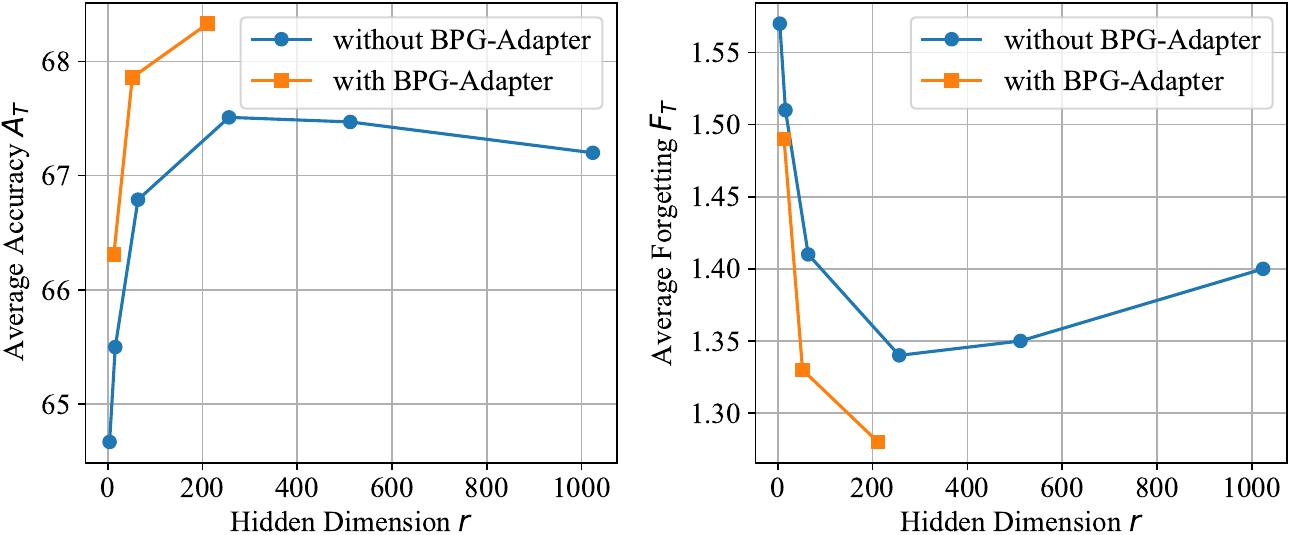}
  \vspace{-2mm}
  \caption{Ablation study on the hidden dimension $r$ and BPG-Adapter. The two curves compare models with and without BPG-Adapter under varying hidden dimensions, reporting both average accuracy $A_T$ (left) and average forgetting $F_T$ (right) on DomainNet.}
  \label{fig:ab_r}
\end{figure}

\paragraph{Effect of BPG-Adapter and Hidden Dimension}
To evaluate the impact of the adapter's hidden dimension on model performance, we conducted experiments on the DomainNet dataset, as illustrated in \cref{fig:ab_r}. For models without BPG-Adapter, a standard adapter with a uniform hidden dimension is assigned to each domain. We set the hidden dimension $r$ to 4, 16, 64, 256, 512, and 1024, measuring both average accuracy and average forgetting. For models equipped with BPG-Adapter, we set the base hidden dimension $r_{0}$ to 4, 16, and 64. Since BPG-Adapter dynamically determines the actual hidden dimension for each domain according to feature separability, the resulting dimension is generally larger than $r_{0}$. To enable a fair comparison, we report the equivalent hidden dimension in this figure. The results reveal two key findings. First, BPG-Adapter consistently outperforms the uniform counterpart in both accuracy and forgetting across comparable hidden dimensions, confirming that adaptively allocating capacity according to domain complexity is more effective than assigning a fixed size to all domains. Second, with BPG-Adapter, a small base dimension ($r_{0}=64$, corresponding to an equivalent dimension around 256) already achieves near-optimal accuracy while maintaining the lowest forgetting, demonstrating that our method enables efficient parameter utilization by assigning larger adapters only to domains that genuinely require additional capacity.

\begin{table}[t]
\footnotesize
\setlength{\tabcolsep}{4pt}
\centering
\caption{Ablation on different fine-tuning methods based on ViT. BPG-A: BPG-Adapter; BPG-I: BPG-Inference.}
\label{tab:tuning}
\vspace{-2mm}
\begin{tabular}{ccccccc}
\toprule
\multirow{2}{*}{Dataset} & \multirow{2}{*}{BPG-A} & \multirow{2}{*}{BPG-I} & \multicolumn{2}{c}{Adapter} & \multicolumn{2}{c}{LoRA} \\
\cmidrule(lr){4-5}\cmidrule(lr){6-7}
 & & & $A_{T}$ ($\uparrow$) & $F_{T}$ ($\downarrow$) & $A_{T}$ ($\uparrow$) & $F_{T}$ ($\downarrow$) \\
\midrule
\multirow{3}{*}{DomainNet} &            &            & 61.69          & 1.86          & 61.69          & 1.86          \\
                           & \ding{51}  &            & 68.33          & 1.28          & 68.04          & 1.39          \\
                           & \ding{51}  & \ding{51}  & \textbf{72.19} & \textbf{0.22} & \textbf{71.83} & \textbf{0.27} \\
\midrule
\multirow{3}{*}{CDDB}      &            &            & 80.52          & 0.93          & 80.52          & 0.93          \\
                           & \ding{51}  &            & 86.94          & 0.85          & 86.92          & 0.88          \\
                           & \ding{51}  & \ding{51}  & \textbf{88.55} & \textbf{0.68} & \textbf{88.48} & \textbf{0.73} \\
\midrule
\multirow{3}{*}{CORe50}    &            &            & 83.23          & --            & 83.23          & --            \\
                           & \ding{51}  &            & 90.15          & --            & 90.36          & --            \\
                           & \ding{51}  & \ding{51}  & \textbf{91.87} & --            & \textbf{92.14} & --            \\
\bottomrule
\end{tabular}
\vspace{-2mm}
\end{table}

\paragraph{Different Tuning Methods}
In existing parameter-isolation DIL methods, two common parameter-efficient fine-tuning approaches are Adapter and LoRA. Our primary experiments are conducted using Adapter. To evaluate the generalizability of our method, we also present experimental results using LoRA in \cref{tab:tuning}. The results demonstrate that the proposed adaptive capacity allocation transfers well to both fine-tuning paradigms. Adapter-based tuning yields slightly stronger results on DomainNet, where domain shifts are more pronounced, whereas LoRA remains competitive on CDDB and CORe50.

\begin{figure*}[t]
  \centering
  \includegraphics[width=0.93\textwidth]{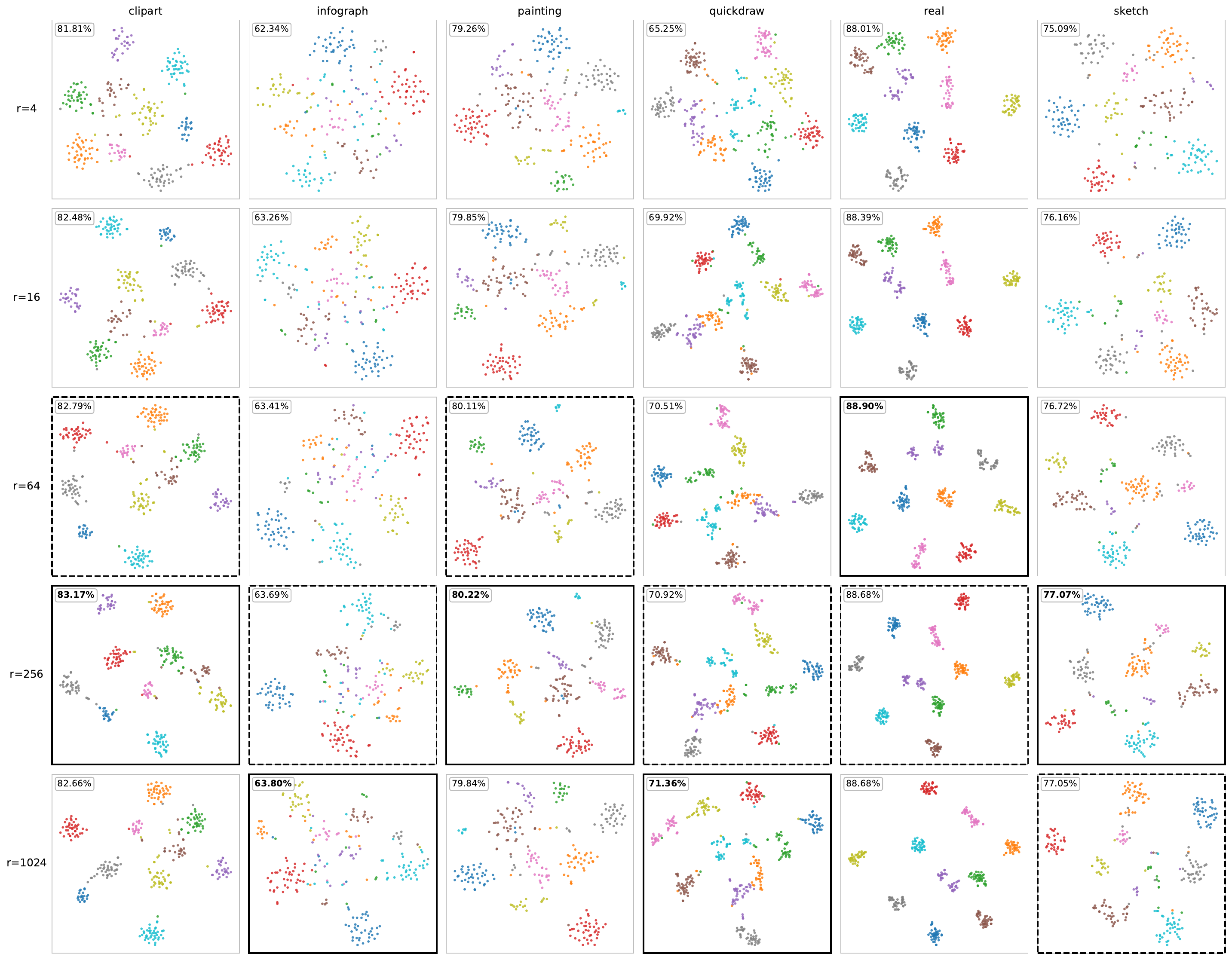}
  \vspace{-2mm}
  \caption{t-SNE visualizations of domain-specific features on DomainNet under uniform adapter hidden dimensions. Each column is a domain and each row a fixed $r$; solid/dashed boxes mark the highest and second-highest accuracy among tested $r$ per column.}
  \label{fig:tsne}
  \vspace{-2mm}
\end{figure*}

\subsection{Additional Analysis on BPG-Inference}
\label{sec:ana_bpg_inference}

\paragraph{Detailed Analysis on BPG-Inference}
\cref{tab:detail_bpg_inference} compares BPG-Adapter with hard domain selection (\emph{BPG-Adapter only}) against the full BPG framework on DomainNet. Hard selection causes Domain~1 accuracy to drop from 78.49\% to 71.06\% as more domains are added, reflecting growing domain-ID misselection. BPG-Inference mitigates this via multi-expert logit fusion. After training on the third domain, the accuracy of Domain~1 improves from 73.64\% to 75.71\% and eventually recovers to 76.83\%.

\begin{table}[t]
\footnotesize
\setlength{\tabcolsep}{1.5pt}
\centering
\caption{Per-domain accuracy matrix on DomainNet: BPG-Adapter only (hard selection) vs.\ full BPG. $D_{1}$ represents Domain~1, and so forth.}
\label{tab:detail_bpg_inference}
\vspace{-2mm}
\begin{tabular}{ccccccc}
\toprule
\multirow{2}{*}{BPG-Adapter only}    & \multicolumn{6}{c}{$A_{T}$ ($\uparrow$): 68.33 $F_{T}$ ($\downarrow$): 1.28} \\ \cmidrule{2-7}
         & Test $D_{1}$ & Test $D_{2}$ & Test $D_{3}$ & Test $D_{4}$ & Test $D_{5}$ & Test $D_{6}$ \\ \midrule
Train $D_{1}$ & 78.49   &         &         &         &         &         \\
Train $D_{2}$ & 74.95   & 43.04   &         &         &         &         \\
Train $D_{3}$ & 73.64   & 42.84   & 67.16   &         &         &         \\
Train $D_{4}$ & 72.81   & 42.43   & 66.85   & 69.31   &         &         \\
Train $D_{5}$ & 71.95   & 41.80   & 66.29   & 69.31   & 78.16   &         \\
Train $D_{6}$ & 71.06   & 40.58   & 65.70   & 69.13   & 77.94   & 63.94   \\ \midrule
\multirow{2}{*}{BPG (full)} & \multicolumn{6}{c}{$A_{T}$ ($\uparrow$): 72.19 $F_{T}$ ($\downarrow$): 0.22} \\ \cmidrule{2-7}
         & Test $D_{1}$ & Test $D_{2}$ & Test $D_{3}$ & Test $D_{4}$ & Test $D_{5}$ & Test $D_{6}$ \\ \midrule
Train $D_{1}$ & 78.49   &         &         &         &         &         \\
Train $D_{2}$ & 74.95   & 43.04   &         &         &         &         \\
Train $D_{3}$ & 75.71   & 42.99   & 70.44   &         &         &         \\
Train $D_{4}$ & 76.09   & 43.50   & 71.56   & 70.82   &         &         \\
Train $D_{5}$ & 76.31   & 44.04   & 71.07   & 70.81   & 82.92   &         \\
Train $D_{6}$ & 76.83   & 44.10   & 70.85   & 70.57   & 83.03   & 68.31   \\
\bottomrule
\end{tabular}
\vspace{-2mm}
\end{table}

\paragraph{Ablation on Weighting Methods in BPG-Inference}
We further ablate logit fusion on the BPG-Adapter-only baseline (\cref{tab:ab_bpg_inference_weight}). Naive averaging (Mean) and max pooling (Max) underperform hard selection, whereas threshold-filtered fusion with L1 normalization (Thres+L1, our default) improves $A_T$ by 3.86\% and reduces $F_T$ to 0.22\%, confirming that ambiguous samples benefit from selective multi-domain aggregation.

\begin{table}[t]
\footnotesize
\setlength{\tabcolsep}{1.8pt}
\centering
\caption{Ablation on BPG-Inference weighting. Hard selection: BPG-Adapter only; Mean: simple average of all logits; Max: select the highest logit value across all domains as the class prediction; Thres: confidence thresholding; L1/L2: weight normalization.}
\label{tab:ab_bpg_inference_weight}
\vspace{-2mm}
\begin{tabular}{c|c|ccc
>{\columncolor[HTML]{E0E0E0}}c
>{\columncolor[HTML]{E0E0E0}}c }
\toprule
                       & Hard selection & Mean   & Max   & Thres+Mean & Thres+L1 & Thres+L2 \\
\midrule
$A_{T}$ ($\uparrow$)   & 68.33  & 56.87  & 66.67 & 71.25      & 72.19    & 72.24    \\
$F_{T}$ ($\downarrow$) & 1.28   & N/A    & 2.08  & 0.49       & 0.22     & 0.26     \\
\midrule
$\Delta A_{T}$         & 0      & -11.46 & -1.66 & +2.92      & +3.86    & +3.91    \\
\bottomrule
\end{tabular}
\vspace{-2mm}
\end{table}

\subsection{Visualization}\label{sec:visualization}
\cref{fig:tsne} complements the ablation in~\cref{fig:ab_r} with t-SNE visualizations of learned features on DomainNet, varying a uniform adapter hidden dimension $r \in \{4, 16, 64, 256, 1024\}$ across all domains (test accuracy annotated per subplot). Domain-wise optima are heterogeneous: easier domains such as Real ($s_5{=}0.530$) peak at smaller $r$ and degrade when given excess capacity, whereas the hardest domains, Infograph ($s_2{=}0.124$) and Quickdraw ($s_4{=}0.151$), continue to improve up to $r{=}1024$; the remaining domains reach their best accuracy at intermediate values (e.g., $r{=}256$). Crucially, uniformly increasing $r$ is not globally optimal. Although hard domains benefit from larger capacity, the same setting increasingly over-parameterizes easier domains and more severely impairs performance on the others, as also evidenced by the rising forgetting in~\cref{fig:ab_r}. This per-domain asymmetry motivates difficulty-aware capacity allocation in BPG-Adapter.

\subsection{Efficiency Analysis}\label{sec:efficiency}
We analyze parameter budget, training overhead, and inference cost on DomainNet (ViT-B/16, $L{=}12$, $d{=}768$, $C{=}345$, $T{=}6$). Per domain, trainable parameters comprise adapters ($2Ld r_t{+}Ld{+}Lr_t$), prompt tokens ($Md$), and a linear classifier ($Cd{+}C$). BPG-Adapter reallocates capacity via~\cref{eq:rt_rule} instead of fixing $r_t{=}r$.

\paragraph{Parameter Budget}
\cref{tab:efficiency_params} reports per-domain $s_t$, $r_t$, and trainable parameter counts. At $r_0{=}64$, harder domains receive larger adapters (e.g., Infograph: $r_2{=}390$) and easier ones smaller capacities (e.g., Real: $r_5{=}91$). Total capacity ($\sum_t r_t{=}1495$, $\bar{r}{\approx}249$) matches a uniform adapter at $r{=}249$ (\cref{fig:ab_r}). With 29.3\,M parameters, BPG surpasses a uniform $r{=}64$ baseline (8.8\,M; $A_T{=}66.79\%$, \cref{fig:ab_r}) at $A_T{=}68.33\%$. Matched to uniform $r{=}256$ (30.0\,M), it still wins on both $A_T$ (68.33\% vs.\ 67.51\%) and $F_T$ (1.28\% vs.\ 1.34\%), showing gains from difficulty-aware reallocation over uniform scaling.

\begin{table}[t]
\setlength{\tabcolsep}{3pt}
\centering
\caption{Parameter budget on DomainNet (ViT-B/16, $r_0=64$). Adapter params.\ are computed as $2 L d r_t + L d + L r_t$ with $L=12$ and $d=768$. Per-domain total includes adapters, prompt tokens, and the linear classifier.}
\label{tab:efficiency_params}
\vspace{-2mm}
\begin{tabular}{lcccc}
\toprule
Domain $t$ & $s_t$ & $r_t$ & Adapter (M) & Domain total (M) \\
\midrule
1 & 0.221 & 219 & 4.05 & 4.32 \\
2 & 0.124 & 390 & 7.20 & 7.48 \\
3 & 0.293 & 165 & 3.05 & 3.33 \\
4 & 0.151 & 320 & 5.91 & 6.18 \\
5 & 0.530 & 91  & 1.69 & 1.96 \\
6 & 0.156 & 310 & 5.73 & 6.00 \\
\midrule
\multicolumn{3}{l}{BPG (adaptive, $r_0=64$)} & 27.63 & \textbf{29.26} \\
\multicolumn{3}{l}{Uniform ($r=64$)}          & 7.14  & 8.77 \\
\multicolumn{3}{l}{Uniform ($r=256$)}         & 28.39 & 30.02 \\
\bottomrule
\end{tabular}
\vspace{-2mm}
\end{table}

\paragraph{Training Overhead}
BPG follows the same sequential protocol: each session updates only the current domain's prompt, adapter, and classifier for 30 epochs. \cref{tab:efficiency_train} reports wall-clock time on DomainNet (ViT-B/16). All methods share a one-time prototype cost of 0.34\,h ($k$-means, $k{=}5$); BPG adds 0.38\,h for separability scoring (one frozen-backbone forward pass per new domain), less than a single training epoch. Larger adapters on harder domains slightly extend main training (22.30\,h vs.\ 21.62\,h at $r{=}64$ and 21.83\,h at $r{=}256$). Overall, BPG finishes in 23.02\,h, only +4.8\% over uniform $r{=}64$ (21.96\,h) and +3.8\% over $r{=}256$ (22.17\,h). Crucially, BPG introduces no learnable router and therefore incurs no joint router-training overhead.

\begin{table}[t]
\footnotesize
\setlength{\tabcolsep}{4pt}
\centering
\caption{Wall-clock training time on DomainNet (ViT-B/16, 30 epochs per domain, $T=6$), in hours on a single GPU aggregated over incremental sessions. Training: main epoch-wise optimization; Separability: one-time feature separability scoring; Prototypes: one-time domain prototype building via $k$-means.}
\label{tab:efficiency_train}
\vspace{-2mm}
\begin{tabular}{lcccc}
\toprule
Method & Training & Separability & Prototypes & Total \\
\midrule
Uniform ($r=64$)           & 21.62 & 0 & 0.34 & 21.96 \\
Uniform ($r=256$)          & 21.83 & 0 & 0.34 & 22.17 \\
BPG (adaptive, $r_0=64$)   & 22.30 & 0.38 & 0.34  & 23.02 \\
\bottomrule
\end{tabular}
\vspace{-2mm}
\end{table}

\paragraph{Inference Latency}
BPG-Inference fuses logits from all $T$ domain-specific models at test time; hard selection uses only the highest-confidence expert. On DomainNet (ViT-B/16, $T{=}6$, single GPU), average per-image latency increases from 11.5\,ms to 74.2\,ms. The overhead is confined to inference and does not alter the training schedule (\cref{tab:efficiency_train}). For offline DIL deployments that prioritize robustness over throughput, 74.2\,ms per image remains modest. Unlike MoE-based soft routing, BPG achieves comparable logit aggregation without a learnable gating module or joint router training. When peak throughput is required, hard selection provides a simple latency-accuracy knob within the same framework.

\section{Conclusion}
We address two key bottlenecks in domain incremental learning: insufficient plasticity from uniform adapter capacity and brittle generalization from hard domain selection at inference. We propose BPG, a unified framework comprising BPG-Adapter, which adaptively allocates adapter hidden dimensions from feature separability, and BPG-Inference, which soft-mixes logits from multiple domain-specific models. Experiments on DomainNet, CDDB, and CORe50 show that BPG achieves state-of-the-art average accuracy with near-zero forgetting, validating the need to jointly adapt plasticity and generalization in lifelong learning.

\paragraph*{Limitations and Future Work}
BPG operates within the parameter-isolation paradigm for rehearsal-free DIL, learning domain-specific parameters sequentially without storing past samples. When a rehearsal buffer is available and privacy constraints permit, combining BPG with selective replay may yield further gains. BPG-Inference increases per-image latency by aggregating logits from all domain experts; this overhead is confined to inference, requires no learnable router, and hard domain selection remains available when peak throughput is critical.


\bibliographystyle{IEEEtran}
\bibliography{main}

\end{document}